\documentclass{article}

\usepackage{times}
\usepackage{graphicx} 

\usepackage{natbib}

\usepackage{algorithm}
\usepackage{algorithmic}
\usepackage{color}
\usepackage{amsmath}
\usepackage{amsfonts}
\usepackage{epstopdf}
\usepackage{caption}
\usepackage{subcaption} 
\usepackage{tikz}
\usetikzlibrary{arrows}
\usepackage{verbatim}
\usepackage{hhline}
\usepackage{setspace}
\usepackage{url}
\usepackage{bm}

\usepackage[arxiv,nohyperref]{icml2016}

\usepackage{array}
\newcolumntype{L}[1]{>{\raggedright\let\newline\\\arraybackslash\hspace{0pt}}m{#1}}
\newcolumntype{C}[1]{>{\centering\let\newline\\\arraybackslash\hspace{0pt}}m{#1}}
\newcolumntype{R}[1]{>{\raggedleft\let\newline\\\arraybackslash\hspace{0pt}}m{#1}}

\newcommand{\nwc}{\newcommand}
\nwc{\as}{\textrm{a.s.}}
\nwc{\defas}{:=}

\nwc{\dist}{\ \sim\ }
\nwc{\distiid}{\stackrel{\mathrm{iid}}{\sim}}

\newcommand{\uu}{\bm{u}}
\newcommand{\U}{\mathcal{U}}

\newcommand{\x}{\bm{x}}
\newcommand{\X}{\mathcal{X}}

\newcommand{\mmu}{\bm{\mu}}
\newcommand{\f}{\bm{f}}
\newcommand{\F}{\bm{F}}
\newcommand{\kk}{\bm{k}}
\newcommand{\K}{\bm{K}}
\newcommand{\PP}{\bm{P}}

\newcommand{\LL}{\bm{L}}

\icmltitlerunning{Continuous Deep Q-Learning with Model-based Acceleration}

\begin{document}

\twocolumn[
\icmltitle{Continuous Deep Q-Learning with Model-based Acceleration}

\icmlauthor{Shixiang Gu$^1$ $^2$ $^3$}{sg717@cam.ac.uk}
\icmlauthor{Timothy Lillicrap$^4$}{countzero@google.com}
\icmlauthor{Ilya Sutskever$^3$}{ilyasu@google.com}
\icmlauthor{Sergey Levine$^3$}{slevine@google.com}
\icmladdress{$^1$University of Cambridge   $^2$Max Planck Institute for Intelligent Systems $^3$Google Brain $^4$Google DeepMind  }

\icmlkeywords{reinforcement learning, deep learning, model learning, actor-critic}

\vskip 0.3in
]

\begin{abstract}
Model-free reinforcement learning has been successfully applied to a range of challenging problems, and has recently been extended to handle large neural network policies and value functions. However, the sample complexity of model-free algorithms, particularly when using high-dimensional function approximators, tends to limit their applicability to physical systems. In this paper, we explore algorithms and representations to reduce the sample complexity of deep reinforcement learning for continuous control tasks. We propose two complementary techniques for improving the efficiency of such algorithms. First, we derive a continuous variant of the Q-learning algorithm, which we call normalized adantage functions (NAF), as an alternative to the more commonly used policy gradient and actor-critic methods. NAF representation allows us to apply Q-learning with experience replay to continuous tasks, and substantially improves performance on a set of simulated robotic control tasks. To further improve the efficiency of our approach, we explore the use of learned models for accelerating model-free reinforcement learning. We show that iteratively refitted local linear models are especially effective for this, and demonstrate substantially faster learning on domains where such models are applicable.
\end{abstract}

\section{Introduction}
\label{sec:intro}

Model-free reinforcement learning (RL) has been successfully applied to a range of challenging problems \cite{kober2012reinforcement,deisenroth2013survey}, and has recently been extended to handle large neural network policies and value functions \cite{mnih2015human,lillicrap2015continuous,wang2015dueling,heess2015learning,hausknecht2015deep,SchulmanLAJM15}. This makes it possible to train policies for complex tasks with minimal feature and policy engineering, using the raw state representation directly as input to the neural network. However, the sample complexity of model-free algorithms, particularly when using very high-dimensional function approximators, tends to be high~\cite{SchulmanLAJM15}, which means that the benefit of reduced manual engineering and greater generality is not felt in real-world domains where experience must be collected on real physical systems, such as robots and autonomous vehicles. In such domains, the methods of choice have been efficient model-free algorithms that use more suitable, task-specific representations \cite{peters2010relative,deisenroth2013survey}, as well as model-based algorithms that learn a model of the system with supervised learning and optimize a policy under this model~\cite{deisenroth2011pilco,levine2015end}. Using task-specific representations dramatically improves efficiency, but limits the range of tasks that can be learned and requires greater domain knowledge. Using model-based RL also improves efficiency, but limits the policy to only be as good as the learned model. For many real-world tasks, it may be easier to represent a good policy than to learn a good model. For example, a simple robotic grasping behavior might only require closing the fingers at the right moment, while the corresponding dynamics model requires learning the complexities of rigid and deformable bodies undergoing frictional contact. It is therefore desirable to bring the generality of model-free deep reinforcement learning into real-world domains by reducing their sample complexity.


In this paper, we propose two complementary techniques for improving the efficiency of deep reinforcement learning in continuous control domains: we derive a variant of Q-learning that can be used in continuous domains, and we propose a method for combining this continuous Q-learning algorithm with learned models so as to accelerate learning while preserving the benefits of model-free RL. Model-free reinforcement learning in domains with continuous actions is typically handled with policy search methods \cite{peters2006policy,peters2010relative}. Integrating value function estimation into these techniques results in actor-critic algorithms \cite{hafner2011reinforcement,lillicrap2015continuous,SchulmanLAJM15},
 which combine the benefits of policy search and value function estimation, but at the cost of training two separate function approximators. Our proposed Q-learning algorithm for continuous domains, which we call normalized advantage functions (NAF), avoids the need for a second actor or policy function, resulting in a simpler algorithm.  
The simpler optimization objective and the choice of value function parameterization result in an algorithm that is substantially more sample-efficient when used with large neural network function approximators on a range of continuous control domains.
 
Beyond deriving an improved model-free deep reinforcement learning algorithm, we also seek to incorporate elements of model-based RL to accelerate learning, without giving up the strengths of model-free methods. One approach is for off-policy algorithms such as Q-learning to incorporate off-policy experience produced by a model-based planner. However, while this solution is a natural one, our empirical evaluation shows that it is ineffective at accelerating learning. As we discuss in our evaluation, this is due in part to the nature of value function estimation algorithms, which must experience both good and bad state transitions to accurately model the value function landscape. We propose an alternative approach to incorporating learned models into our continuous-action Q-learning algorithm based on \emph{imagination rollouts}: on-policy samples generated under the learned model, analogous to the Dyna-Q method \cite{sutton1990integrated}. We show that this is extremely effective when the learned dynamics model perfectly matches the true one, but degrades dramatically with imperfect learned models. However, we demonstrate that iteratively fitting local linear models to the latest batch of on-policy or off-policy rollouts provides sufficient \emph{local} accuracy to achieve substantial improvement using short imagination rollouts in the vicinity of the real-world samples.

Our paper provides three main contributions: first, we derive and evaluate a Q-function representation that allows for effective Q-learning in continuous domains. Second, we evaluate several na\"{i}ve options for incorporating learned models into model-free Q-learning, and we show that they are minimally effective on our continuous control tasks. Third, we propose to combine locally linear models with local on-policy imagination rollouts to accelerate model-free continuous Q-learning, and show that this produces a large improvement in sample complexity. We evaluate our method on a series of simulated robotic tasks and compare to prior methods.




\section{Related Work}
\label{sec:related}

Deep reinforcement learning has received considerable attention in recent years due to its potential to automate the design of representations in RL. Deep reinforcement learning and related methods have been applied to learn policies to play Atari games \cite{mnih2015human,schaul2015prioritized} and perform a wide variety of simulated and real-world robotic control tasks \cite{hafner2011reinforcement,lillicrap2015continuous,levine2013guided,deimportance,hafner2011reinforcement}. While the majority of deep reinforcement learning methods in domains with discrete actions, such as Atari games, are based around value function estimation and Q-learning \cite{mnih2015human}, continuous domains typically require explicit representation of the policy, for example in the context of a policy gradient algorithm \cite{SchulmanLAJM15}. If we wish to incorporate the benefits of value function estimation into continuous deep reinforcement learning, we must typically use two networks: one to represent the policy, and one to represent the value function \cite{SchulmanLAJM15,lillicrap2015continuous}. In this paper, we instead describe how the simplicity and elegance of Q-learning can be ported into continuous domains, by learning a single network that outputs both the value function and policy. Our Q-function representation is related to dueling networks \cite{wang2015dueling}, though our approach applies to continuous action domains. Our empirical evaluation demonstrates that our continuous Q-learning algorithm achieves faster and more effective learning on a set of benchmark tasks compared to continuous actor-critic methods, and we believe that the simplicity of this approach will make it easier to adopt in practice. Our Q-learning method is also related to the work of \citet{rawlik2013stochastic}, but the form of our Q-function update is more standard. 

As in standard RL, model-based deep reinforcement learning methods have generally been more efficient \cite{li2004iterative,watter2015embed,li2004iterative,wahlstrom2015pixels,levine2013guided}, while model-free algorithms tend to be more generally applicable but substantially slower \cite{SchulmanLAJM15,lillicrap2015continuous}. Combining model-based and model-free learning has been explored in several ways in the literature. The method closest to our imagination rollouts approach is Dyna-Q \cite{sutton1990integrated}, which uses simulated experience in a learned model to supplement real-world on-policy rollouts. As we show in our evaluation, using Dyna-Q style methods to accelerate model-free RL is very effective when the learned model perfectly matches the true model, but degrades rapidly as the model becomes worse. We demonstrate that using iteratively refitted local linear models achieves substantially better results with imagination rollouts than more complex neural network models. We hypothesize that this is likely due to the fact that the more expressive models themselves require substantially more data, and that otherwise efficient algorithms like Dyna-Q are vulnerable to poor model approximations.




\section{Background}
\label{sec:bg}



In reinforcement learning, the goal is to learn a policy to control a system with states $\x \in \X$ and actions $\uu \in \U$ in environment $E$, so as to maximize the expected sum of returns according to a reward function $r(\x,\uu)$. The dynamical system is defined by an initial state distribution $p(\x_1)$ and a dynamics distribution $p(\x_{t+1}|\x_t, \uu_t)$. At each time step $t \in [1,T]$, the agent chooses an action $\uu_t$ according to its current policy $\pi(\uu_t | \x_t)$, and observes a reward $r(\x_t, \uu_t)$. The agent then experiences a transition to a new state sampled from the dynamics distribution, and we can express the resulting state visitation frequency of the policy $\pi$ as $\rho^\pi(\x_t)$. Define $R_t=\sum_{i=t}^T\gamma^{(i-t)}r(\x_i,\uu_i)$, the goal is to maximize the expected sum of returns, given by $R=\mathbb{E}_{r_{i \geq 1}, \x_{i \geq 1} \sim E, \uu_{i \geq 1}\sim \pi}[R_1]$, where $\gamma$ is a discount factor that prioritizes earlier rewards over later ones. With $\gamma < 1$, we can also set $T = \infty$, though we use a finite horizon for all of the tasks in our experiments. The expected return $R$ can be optimized using a variety of model-free and model-based algorithms. In this section, we review several of these methods that we build on in our work.




\paragraph{Model-Free Reinforcement Learning.} When the system dynamics $p(\x_{t+1}|\x_t,\uu_t)$ are not known, as is often the case with physical systems such as robots, policy gradient methods~\cite{peters2006policy} and value function or Q-function learning with function approximation~\cite{sutton1999policy} are often preferred. Policy gradient methods provide a simple, direct approach to RL, which can succeed on high-dimensional problems, but potentially requires a large number of samples~\cite{SchulmanLAJM15,schulman2015high}. Off-policy algorithms that use value or Q-function approximation can in principle achieve better data efficiency~\cite{lillicrap2015continuous}. However, adapting such methods to continuous tasks typically requires optimizing two function approximators on different objectives. We instead build on standard Q-learning, which has a single objective. We summarize Q-learning in this section. The Q function $Q^\pi(\x_t, \uu_t)$ corresponding to a policy $\pi$ is defined as the expected return from $\x_t$ after taking action $\uu_t$ and following the policy $\pi$ thereafter:
\begin{equation} \label{eq:q}
\begin{split}
Q^\pi(\x_t, \uu_t) = \mathbb{E}_{r_{i \geq t}, \x_{i > t} \sim E, \uu_{i > t}\sim \pi}[R_t|\x_t,\uu_t] \\
\end{split}
\end{equation}  
Q-learning learns a greedy deterministic policy \mbox{$\mmu(\x_t) = \arg\max_{\uu} Q(\x_t,\uu_t)$}, which corresponds to \mbox{$\pi(\uu_t|\x_t) = \delta(\uu_t=\mmu(\x_t))$}. Let $\theta^Q$ parametrize the action-value function, and $\beta$ be an arbitrary exploration policy, the learning objective is to minimize the Bellman error, where we fix the target $y_t$:
\begin{equation} \label{eq:qlearn}
\begin{split}
L(\theta^Q) &= \mathbb{E}_{\x_t\sim\rho^\beta,\uu_t\sim\beta,r_t\sim E}[(Q(\x_t,\uu_t|\theta^Q)-y_t)^2] \\
y_t &= r(\x_t,\uu_t) + \gamma Q(\x_{t+1},\mmu(\x_{t+1})) 
\end{split}
\end{equation}  
For continuous action problems, Q-learning becomes difficult, because it requires maximizing a complex, nonlinear function at each update. For this reason, continuous domains are often tackled using actor-critic methods~\cite{konda1999actor,hafner2011reinforcement,silver2014deterministic,lillicrap2015continuous}, where a separate parameterized ``actor" policy $\pi$ is learned in addition to the Q-function or value function ``critic,'' such as Deep Deterministic Policy Gradient (DDPG) algorithm~\cite{lillicrap2015continuous}. 


In order to describe our method in the following sections, it will be useful to also define the value function $V^\pi(\x_t,\uu_t)$ and advantage function $A^\pi(\x_t, \uu_t)$ of a given policy $\pi$:
\begin{equation} \label{eq:va}
\begin{split}
& V^\pi(\x_t) = \mathbb{E}_{r_{i \geq t}, \x_{i > t} \sim E, \uu_{i \geq t}\sim \pi}[R_t|\x_t,\uu_t] \\
& A^\pi(\x_t, \uu_t) = Q^\pi(\x_t, \uu_t) - V^\pi(\x_t).
\end{split}
\end{equation}

\paragraph{Model-Based Reinforcement Learning.}
If we know the dynamics $p(\x_{t+1}|\x_t,\uu_t)$, or if we can approximate them with some learned model $\hat{p}(\x_{t+1}|\x_t,\uu_t)$, we can use model-based RL and optimal control. While a wide range of model-based RL and control methods have been proposed in the literature \cite{deisenroth2013survey,kober2012reinforcement}, two are particularly relevant for this work: iterative LQG (iLQG) \cite{li2004iterative} and Dyna-Q \cite{sutton1990integrated}.  The iLQG algorithm optimizes trajectories by iteratively constructing locally optimal linear feedback controllers under a local linearization of the dynamics \mbox{$\hat{p}(\x_{t+1}|\x_t,\uu_t)=\mathcal{N}(\f_{\x t}\x_t + \f_{\uu t}\uu_t, \F_t)$} and a quadratic expansion of the rewards $r(\x_t,\uu_t)$ ~\cite{tassa2012synthesis}. Under linear dynamics and quadratic rewards, the action-value function $Q(\x_t, \uu_t)$ and value function $V(\x_t)$ are locally quadratic and can be computed by dynamics programming. The optimal policy can be derived analytically from the quadratic $Q(\x_t, \uu_t)$ and $V(\x_t)$ functions, and corresponds to a linear feedback controller \mbox{$\bm{g}(\x_t) = \hat{\uu}_t + \kk_t + \K_t(\x_t - \hat{\x}_t)$}, where $\kk_t$ is an open-loop term, $\K_t$ is the closed-loop feedback matrix, and $\hat{\x}_t$ and $\hat{\uu}_t$ are the states and actions of the nominal trajectory, which is the average trajectory of the controller. Employing the maximum entropy objective~\cite{levine2013guided}, we can also construct a linear-Gaussian controller, where $c$ is a scalar to adjust for arbitrary scaling of the reward magnitudes:
  \begin{equation} \label{eq:ilqg_controller}
  \begin{split}
\pi^{iLQG}_t(\uu_t|\x_t) = \mathcal{N}(\hat{\uu}_t + \kk_t + \K_t(\x_t-\hat{\x}_t), -cQ_{\uu,\uu t}^{-1})
  \end{split}
  \end{equation}
  When the dynamics are not known, a particularly effective way to use iLQG is to combine it with learned time-varying linear models $\hat{p}(\x_{t+1}|\x_t,\uu_t)$. In this variant of the algorithm, trajectories are sampled from the controller in Equation~(\ref{eq:ilqg_controller}) and used to fit time-varying linear dynamics with linear regression. These dynamics are then used with iLQG to obtain a new controller, typically using a KL-divergence constraint to enforce a trust region, so that the new controller doesn't deviate too much from the region in which the samples were generated \cite{levine2014learning}.

Besides enabling iLQG and other planning-based algorithms, a learned model of the dynamics can allow a model-free algorithm to generate synthetic experience by performing rollouts in the learned model. A particularly relevant method of this type is Dyna-Q \cite{sutton1990integrated}, which performs real-world rollouts using the policy $\pi$, and then generates synthetic rollouts using a model learned from these samples. The synthetic rollouts originate at states visited by the real-world rollouts, and serve as supplementary data for a variety of possible reinforcement learning algorithms. However, most prior Dyna-Q methods have focused on relatively small, discrete domains. In Section~\ref{sec:fictional}, we describe how our method can be extended into a variant of Dyna-Q to achieve substantially faster learning on a range of continuous control tasks with complex neural network policies, and in Section~\ref{sec:experiments}, we empirically analyze the sensitivity of this method to imperfect learned dynamics models.

\section{Continuous Q-Learning with Normalized Advantage Functions}
\label{sec:normq}
We first propose a simple method to enable Q-learning in continuous action spaces with deep neural networks, which we refer to as normalized advantage functions (NAF). The idea behind normalized advantage functions is to represent the Q-function $Q(\x_t, \uu_t)$ in Q-learning in such a way that its maximum, $\arg\max_{\uu} Q(\x_t, \uu_t)$, can be determined easily and analytically during the Q-learning update. While a number of representations are possible that allow for analytic maximization, the one we use in our implementation is based on a neural network that separately outputs a value function term $V(\x)$ and an advantage term $A(\x,\uu)$, which is parameterized as a quadratic function of nonlinear features of the state:
   \begin{equation} \label{eq:normq}
   \begin{split}
   Q(\x,\uu|\theta^Q) &= A(\x,\uu|\theta^A) + V(\x|\theta^V) \\
   A(\x,\uu|\theta^A) &= -\frac{1}{2}(\uu-\mmu(\x|\theta^\mu))^T \PP(\x|\theta^P) (\uu-\mmu(\x|\theta^\mu))\nonumber\\
   \end{split}
   \end{equation}
   $\PP(\x|\theta^P)$ is a state-dependent, positive-definite square matrix, which is parametrized by $\PP(\x|\theta^P) = \LL(\x|\theta^P)\LL(\x|\theta^P)^T$, where $\LL(\x|\theta^P)$ is a lower-triangular matrix whose entries come from a linear output layer of a neural network, with the diagonal terms exponentiated. While this representation is more restrictive than a general neural network function, since the Q-function is quadratic in $\uu$, the action that maximizes the Q-function is always given by $\mmu(\x|\theta^\mu)$. We use this representation with a deep Q-learning algorithm analogous to \citet{mnih2015human}, using target networks and a replay buffers as described by \cite{lillicrap2015continuous}. NAF, given by Algorithm~\ref{alg:nafq}, is considerably simpler than DDPG.
   \begin{algorithm}
   	\footnotesize
   	\caption{Continuous Q-Learning with NAF}
   	\label{alg:nafq}
   	\begin{algorithmic} 
   		\STATE Randomly initialize normalized Q network $Q(\x,\uu|\theta^Q)$.
   		\STATE Initialize target network $Q'$ with weight $\theta^{Q'}\leftarrow \theta^Q$.
   		\STATE Initialize replay buffer $R\leftarrow\emptyset$.
   		\FOR{episode=$1,M$}
   		\STATE Initialize a random process $\mathcal{N}$ for action exploration
   		\STATE Receive initial observation state $\x_1\sim p(\x_1)$
   		\FOR{t=$1,T$}
   		\STATE Select action $\uu_t=\mu(\x_t|\theta^\mu)+\mathcal{N}_t$
   		\STATE Execute $\uu_t$ and observe $r_t$ and $\x_{t+1}$
   		\STATE Store transition ($\x_t,\uu_t,r_t,\x_{t+1}$) in $R$
   		\FOR{iteration=$1, I$}
   		\STATE Sample a random minibatch of $m$ transitions from $R$
   		\STATE Set $y_i = r_i + \gamma V'(\x_{i+1}|\theta^{Q'})$
   		\STATE Update $\theta^Q$ by minimizing the loss: $L=\frac{1}{N}\sum_i (y_i - Q(\x_i, \uu_i|\theta^Q))^2$
   		\STATE Update the target network: $\theta^{Q'}\leftarrow \tau\theta^Q + (1-\tau)\theta^{Q'}$
   		\ENDFOR
   		\ENDFOR
   		\ENDFOR
   	\end{algorithmic}
   \end{algorithm}

Decomposing $Q$ into an advantage term $A$ and a state-value term $V$ was suggested by ~\citet{baird1993advantage,harmon1996multi}, and was recently explored by~\citet{wang2015dueling} for discrete action problems. Normalized action-value functions have also been proposed by ~\citet{rawlik2013stochastic} in the context of an alternative temporal difference learning algorithm. However, our method is the first to combine such representations with deep neural networks into an algorithm that can be used to learn policies for a range of challenging continuous control tasks. In general, $A$ does not need to be quadratic, and exploring other parametric forms such as multimodal distributions is an interesting avenue for future work. The appendix provides details on adaptive exploration rule derivation with experimental results, and a variational interpretation of Q-learning which gives an intuitive explanation of the behavior of NAF that conforms with empirical results.
 
\section{Accelerating Learning with Imagination Rollouts}
\label{sec:fictional}


While NAF provides some advantages over actor-critic model-free RL methods in continuous domains, we can improve their data efficiency substantially under some additional assumptions by exploiting learned models. We will show that incorporating a particular type of learned model into Q-learning with NAFs significantly improves sample efficiency, while still allowing the final policy to be finetuned with model-free learning to achieve good performance without the limitations of imperfect models.


\subsection{Model-Guided Exploration}
\label{sec:ilqg_guide}
One natural approach to incorporating a learned model into an off-policy algorithm such as Q-learning is to use the learned model to generate good exploratory behaviors using planning or trajectory optimization. To evalaute this idea, we utilize the iLQG algorithm to generate good trajectories under the model, and then mix these trajectories together with on-policy experience by appending them to the replay buffer. Interestingly, we show in our evaluation that, even when planning under the true model, the improvement obtained from this approach is often quite small, and varies significantly across domains and choices of exploration noise. The intuition behind this result is that off-policy iLQG exploration is too different from the learned policy, and Q-learning must consider alternatives in order to ascertain the optimality of a given action. That is, it's not enough to simply show the algorithm \emph{good} actions, it must also experience bad actions to understand which actions are better and which are worse.


\subsection{Imagination Rollouts}
As discussed in the previous section, incorporating off-policy exploration from good, narrow distributions, such as those induced by iLQG, often does not result in significant improvement for Q-learning.
These results suggest that Q-learning, which learns a policy based on minimizing temporal differences, inherently requires noisy on-policy actions to succeed. In real-world domains such as robots and autonomous vehicles, this can be undesirable for two reasons: first, it suggests that large amounts of on-policy experience are required in addition to good off-policy samples, and second, it implies that the policy must be allowed to make ``its own mistakes'' during training, which might involve taking undesirable or dangerous actions that can damage real-world hardware.

One way to avoid these problems while still allowing for a large amount of on-policy exploration is to generate synthetic on-policy trajectories under a learned model. Adding these synthetic samples, which we refer to as \textit{imagination rollouts}, to the replay buffer effectively augments the amount of experience available for Q-learning. The particular approach we use is to perform rollouts in the real world using a mixture of planned iLQG trajectories and on-policy trajectories, with various mixing coefficients evaluated in our experiments,
 and then generate additional synthetic on-policy rollouts using the learned model from each state visited along the real-world rollouts. This approach can be viewed as a variant of the Dyna-Q algorithm~\cite{sutton1990integrated}. However, while Dyna-Q has primarily been used with small and discrete systems, we show that using iteratively refitted linear models allows us to extend the approach to deep reinforcement learning on a range of continuous control domains. In some scenarios, we can even generate all or most of the real rollouts using off-policy iLQG controllers, which is desirable in safety-critic domains where poorly trained policies might take dangerous actions. The algorithm is given as Algorithm~\ref{alg:dynaq}, and is an extension on Algorithm~\ref{alg:nafq} combining model-based RL.
\begin{algorithm}
	\footnotesize
	\caption{Imagination Rollouts with Fitted Dynamics and Optional iLQG Exploration}
	\label{alg:dynaq}
	\begin{algorithmic} 
		\STATE Randomly initialize normalized Q network $Q(\x,\uu|\theta^Q)$.
		\STATE Initialize target network $Q'$ with weight $\theta^{Q'}\leftarrow \theta^Q$.
		\STATE Initialize replay buffer $R\leftarrow\emptyset$ and fictional buffer $R_f\leftarrow\emptyset$.
		\STATE Initialize additional buffers $B\leftarrow\emptyset,B_{old}\leftarrow\emptyset$ with size $nT$.
		\STATE Initialize fitted dynamics model $\mathcal{M}\leftarrow\emptyset$. 
		\FOR{$episode=1, M$}
		\STATE Initialize a random process $\mathcal{N}$ for action exploration
		\STATE Receive initial observation state $\x_1$
		\STATE Select $\mu'(\x, t)$ from \{$\mu(\x|\theta^\mu), \pi^{iLQG}_t(\uu_t|\x_t)$\} with probabilities \{$p,1-p$\}
		\FOR{$t=1,T$}
		\STATE Select action $\uu_t=\mu'(\x_t,t)+\mathcal{N}_t$
		\STATE Execute $\uu_t$ and observe $r_t$ and $\x_{t+1}$
		\STATE Store transition ($\x_t,\uu_t,r_t,\x_{t+1},t$) in $R$ and $B$
		
		\IF{$\mod(episode\cdot T+t,m)=0$ and $\mathcal{M}\neq\emptyset$}
		\STATE Sample $m$ ($\x_i,\uu_i,r_i,\x_{i+1},i$) from $B_{old}$
		\STATE Use $\mathcal{M}$ to simulate $l$ steps from each sample 
		\STATE Store all fictional transitions in $R_f$
		\ENDIF
		\STATE Sample a random minibatch of $m$ transitions $I\cdot l$ times from $R_f$  and $I$ times from $R$, and update $\theta^Q,\theta^{Q'}$ as in Algorithm~\ref{alg:nafq} per minibatch.
		\ENDFOR
		\IF{$B_f$ is full}
		\STATE $\mathcal{M}\leftarrow$ FitLocalLinearDynamics($B_f$) (see Section~\ref{sec:fitting})
		\STATE $\pi^{iLQG}\leftarrow$ iLQG\_OneStep($B_f,\mathcal{M}$) (see appendix)
		\STATE $B_{old}\leftarrow B_f, B_f\leftarrow\emptyset$ 
		\ENDIF
		\ENDFOR
	\end{algorithmic}
\end{algorithm}


Imagination rollouts can suffer from severe bias when the learned model is inaccurate. For example, we found it very difficult to train nonlinear neural network models for the dynamics that would actually improve the efficiency of Q-learning when used for imagination rollouts. As discussed in the following section, we found that using iteratively refitted time-varying linear dynamics produced substantially better results. In either case, we would still like to preserve the generality and optimality of model-free RL while deriving the benefits of model-based learning. To that end, we observe that most of the benefit of model-based learning is derived in the early stages of the learning process, when the policy induced by the neural network Q-function is poor. As the Q-function becomes more accurate, on-policy behavior tends to outperform model-based controllers. We therefore propose to switch off imagination rollouts after a given number of iterations.\footnote{In future work, it would be interesting to select this iteration adaptively based on the expected relative performance of the Q-function policy and model-based planning.} In this framework, the imagination rollouts can be thought of as an inexpensive way to pretrain the Q-function, such that fine-tuning using real world experience can quickly converge to an optimal solution.


\subsection{Fitting the Dynamics Model}
\label{sec:fitting}
In order to obtain good imagination rollouts and improve the efficiency of Q-learning, we needed to use an effective and data-efficient model learning algorithm. While prior methods propose a variety of model classes, including neural networks \cite{heess2015learning}, Gaussian processes \cite{deisenroth2011pilco}, and locally-weighted regression \cite{atkeson1997locally}, we found that we could obtain good results by using iteratively refitted time-varying linear models, as proposed by \citet{levine2014learning}. In this approach, instead of learning a good global model for all states and actions, we aim only to obtain a good local model around the latest set of samples. This approach requires a few additional assumptions: namely, it requires the initial state to be either deterministic or low-variance Gaussian, and it requires the states and actions to all be continuous. To handle domains with more varied initial states, we can use a mixture of Gaussian initial states with separate time-varying linear models for each one. The model itself is given by $p_t(\x_{t+1} | \x_t, \uu_t) = \mathcal{N}(\mathbf{F}_t [\x_t ; \uu_t] + \mathbf{f}_t, \mathbf{N}_t)$. Every $n$ episodes, we refit the parameters $\mathbf{F}_t$, $\mathbf{f}_t$, and $\mathbf{N}_t$ by fitting a Gaussian distribution at each time step to the vectors $[\x_t^i; \uu_t^i; \x_{t+1}^i]$, where $i$ indicates the sample index, and conditioning this Gaussian on $[\x_t;\uu_t]$ to obtain the parameters of the linear-Gaussian dynamics at that step. We use $n = 5$ in our experiments. Although this approach introduces additional assumptions beyond the standard model-free RL setting, we show in our evaluation that it produces impressive gains in sample efficiency on tasks where it can be applied.

\section{Experiments}
\label{sec:experiments}



We evaluated our approach on a set of simulated robotic tasks using the MuJoCo simulator \cite{todorov2012mujoco}. The tasks were based on the benchmarks described by \citet{lillicrap2015continuous}. Although we attempted to replicate the tasks in previous work as closely as possible, discrepancies in the simulator parameters and the contact model produced results that deviate slightly from those reported in prior work. In all experiments, the input to the policy consisted of the state of the system, defined in terms of joint angles and root link positions. Angles were often converted to sine and cosine encoding.

\begin{figure*}[t!]
	\centering
		\begin{subfigure}[t]{0.33\textwidth}
			\centering
			\includegraphics[width=0.9\linewidth]{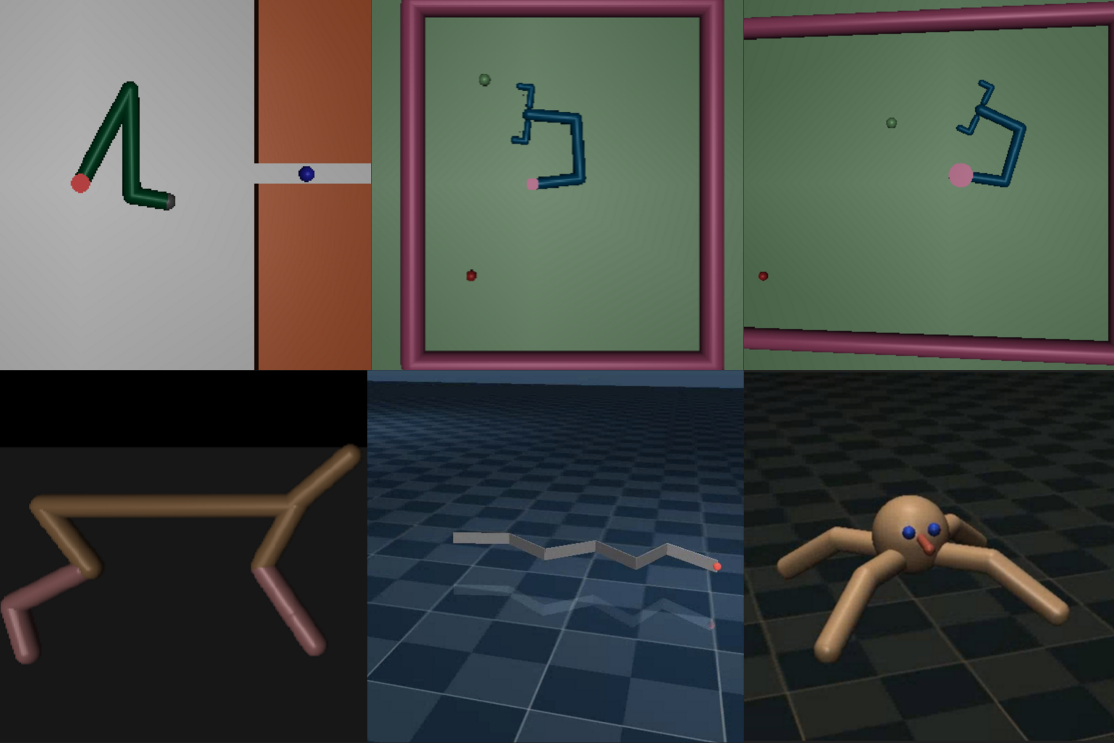}
			\caption{Example task domains.}
			\label{fig:domains}
		\end{subfigure}
	\begin{subfigure}[t]{0.33\textwidth}
		\centering
	\includegraphics[width=0.9\linewidth]{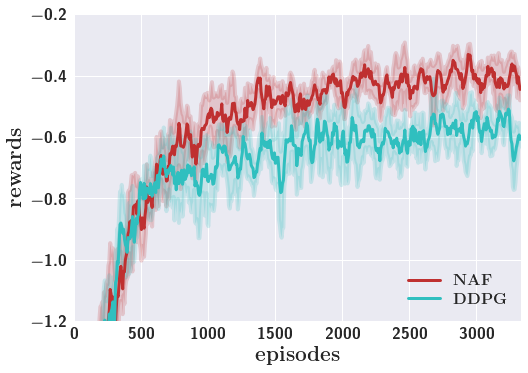}
	\caption{NAF and DDPG on multi-target reacher. }
	\label{fig:normq_reacher}
\end{subfigure}
\begin{subfigure}[t]{0.33\textwidth}
	\centering
	\includegraphics[width=0.9\linewidth]{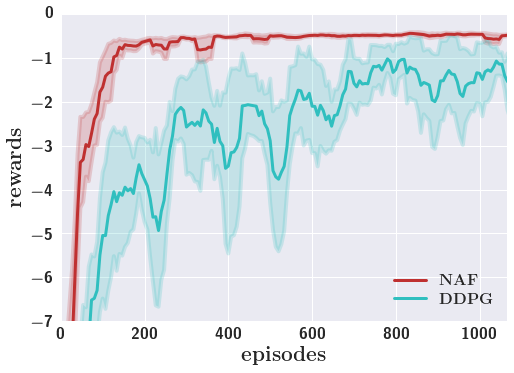}
	\caption{NAF and DDPG on peg insertion.}
	\label{fig:normq_peg}
\end{subfigure}
\caption{(a) Task domains: top row from left (manipulation tasks: peg, gripper, mobile gripper), bottom row from left (locomotion tasks: cheetah, swimmer6, ant). (b,c) NAF vs DDPG results on three-joint reacher and peg insertion. On reacher, the DDPG policy continuously fluctuates the tip around the target, while NAF stabilizes well at the target.\vspace{-0.1 in}}
\label{fig:naf}
\end{figure*}
For both our method and the prior DDPG~\cite{lillicrap2015continuous} algorithm in the comparisons, we used neural networks with two layers of 200 rectified linear units (ReLU) to produce each of the output parameters -- the Q-function and policy in DDPG, and the value function $V$, the advantage matrix $\LL$, and the mean $\mmu$ for NAF. Since Q-learning was done with a replay buffer, we applied the Q-learning update 5 times per each step of experience to accelerate learning ($I=5$). To ensure a fair comparison, DDPG also updates both the Q-function and policy parameters 5 times per step.

\subsection{Normalized Advantage Functions}
\label{sec:exp_naf}
In this section, we compare NAF and DDPG on 10 representative domains from~\citet{lillicrap2015continuous}, with three additional domains: a four-legged 3D ant, a six-joint 2D swimmer, and a 2D peg (see the appendix for the descriptions of task domains). We found the most sensitive hyperparameters to be presence or absence of batch normalization, base learning rate for ADAM~\cite{kingma2014adam} $\in\{1e^{-4},1e^{-3},1e^{-2}\}$, and exploration noise scale $\in\{0.1,0.3,1.0\}$. We report the best performance for each domain. We were unable to achieve good results with the method of \citet{rawlik2013stochastic} on our domains, likely due to the complexity of high-dimensional neural network function approximators.

Figure~\ref{fig:normq_reacher}, Figure~\ref{fig:normq_peg}, and additional figures in the appendix show the performances on the three-joint reacher, peg insertion, and a gripper with mobile base. While the numerical gap in reacher may be small, qualitatively there is also a very noticeable difference between NAF and DDPG. DDPG converges to a solution where the deterministic policy causes the tip to fluctuate continuously around the target, and does not reach it precisely. NAF, on the other hand, learns a smooth policy that makes the tip slow down and stabilize at the target. This difference is more noticeable in peg insertion and moving gripper, as shown by the much faster convergence rate to the optimal solution. Precision is very important in many real-world robotic tasks, and these result suggest that NAF may be preferred in such domains.

On locomotion tasks, the performance of the two methods is relatively similar. On the six-joint swimmer task and four-legged ant, NAF slightly outperforms DDPG in terms of the convergence speed; however, DDPG is faster on cheetah and finds a better policy on walker2d. The loss in performance of NAF can potentially be explained by downside of the mode-seeking behavior as analyzed in the appendix, where it is hard to explore other modes once the quadratic advantage function finds a good one. Choosing a parametric form that is more expressive than a quadratic could be used to address this limitation in future work.

The results on all of the domains are summarized in Table~\ref{tab:normq}. Overall, NAF outperformed DDPG on the majority of tasks, particularly manipulation tasks that require precision and suffer less from the lack of multimodal Q-functions. This makes this approach particularly promising for efficient learning of real-world robotic tasks.

  \begin{table}[ht]
  	\centering 
  	\footnotesize
  	\begin{tabular}{|c  |c |c c| c c  | }
  		\hline
  		Domains	&-	& DDPG & episodes &  NAF& episodes \\
  		\hline
  		Cartpole &-2.1& -0.601 & 420 & -0.604 & \textbf{190} \\ 
  		Reacher &-2.3& -0.509 & 1370 & \textbf{-0.331} & \textbf{1260} \\ 
  		Peg &-11& -0.950 & 690 & \textbf{-0.438} & \textbf{130}  \\ 
  		Gripper& -29& 1.03 & 2420 & \textbf{1.81} & \textbf{1920} \\ 
  		GripperM& -90& -20.2 & 1350 & \textbf{-12.4} & \textbf{730} \\ 
  		Canada2d &-12& -4.64 & 1040& \textbf{-4.21} & 900 \\ 
  		Cheetah &-0.3& \textbf{8.23}  & \textbf{1590} & 7.91 & 2390 \\ 
  		Swimmer6& -325 & -174 & 220 & \textbf{-172} & \textbf{190} \\ 
  		Ant &-4.8& -2.54 & 2450 & -2.58 & \textbf{1350} \\ 
  		Walker2d &0.3& \textbf{2.96} & \textbf{850} & 1.85 & 1530 \\ 
  		\hline
  	\end{tabular}
  	\caption{\footnotesize Best test rewards of DDPG and NAF policies, and the episodes it requires to reach within 5\% of the best value. ``-" denotes scores by a random agent.\vspace{-0.1 in}}
  	\label{tab:normq}
  \end{table}

\subsection{Evaluating Best-Case Model-Based Improvement with True Models}
\label{sec:exp_model_rl}

In order to determine how best to incorporate model-based components to accelerate model-free Q-learning, we tested several approaches using the ground truth dynamics, to control for challenges due to model fitting. We evaluated both of the methods discussed in Section~\ref{sec:fictional}: the use of model-based planning to generate good off-policy rollouts in the real world, and the use of the model to generate on-policy synthetic rollouts.


\begin{figure*}[t!]
\begin{subfigure}[t]{0.33\textwidth}
	\centering\captionsetup{width=.8\linewidth}%
	\includegraphics[width=0.99\linewidth]{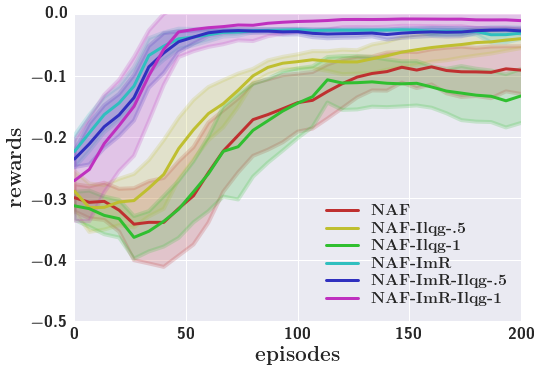}
	\caption{NAF on single-target reacher.  \vspace{-0.1 in}}
	\label{fig:normq_fixedreacher_true}
\end{subfigure}
\begin{subfigure}[t]{0.33\textwidth}
	\centering\captionsetup{width=.8\linewidth}%
		\includegraphics[width=0.99\linewidth]{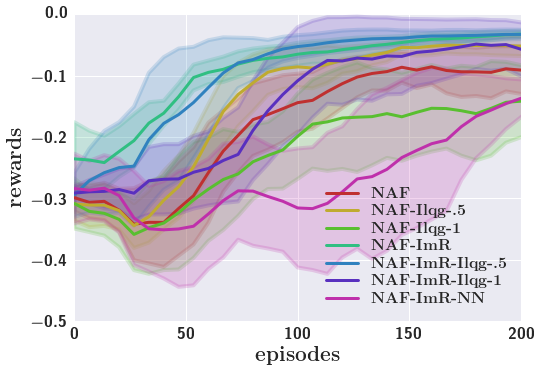}
	\caption{NAF on single-target reacher.  \vspace{-0.1 in}}
	\label{fig:normq_fixedreacher_fitted}
\end{subfigure}
\begin{subfigure}[t]{0.33\textwidth}
	\centering\captionsetup{width=.8\linewidth}%
	\includegraphics[width=0.99\linewidth]{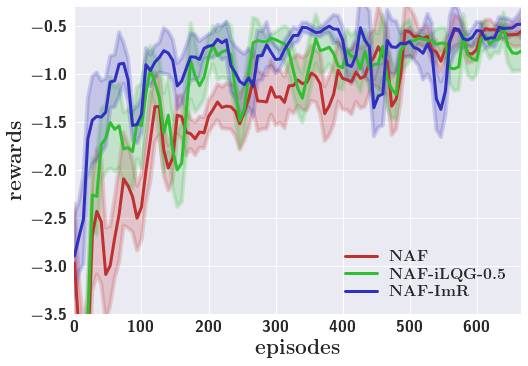}
	\caption{NAF on single-target gripper. \vspace{-0.1 in}}
	\label{fig:normq_fixedgrip_fitted}
\end{subfigure}
\caption{Results on NAF with iLQG-guided exploration and imagination rollouts (a) using true dynamics (b,c) using fitted dynamics. ``ImR" denotes using the imagination rollout with $l=10$ steps on the reacher and $l=5$ steps on the gripper. ``iLQG-$x$" indicates mixing $x$ fraction of iLQG episodes. Fitted dynamics uses time-varying linear models with sample size $n=5$, except ``-NN" which fits a neural network to global dynamics.\vspace{-0.1 in}}
\label{fig:imr}
\end{figure*}
Figure~\ref{fig:normq_fixedreacher_true} shows the effect of mixing off-policy iLQG experience and imagination rollouts on the three-joint reacher. It is noticeable that mixing the good off-policy experience does not significantly improve data-efficiency, while imagination rollouts always improve data-efficiency or final performance significantly. In the context of Q-learning, this result is not entirely surprising: Q learning must experience both good and bad actions in order to determine which actions are preferred, while the good model-based rollouts are so far removed from the policy in the early stages of learning that they provide little useful information. Figure~\ref{fig:normq_fixedreacher_true} also evaluates two different variants of the imagination rollouts approach, where the rollouts in the real world are performed either using the learned policy, or using model-based planning with iLQG. In the case of this task, the iLQG rollouts achieve slightly better results, since the on-policy imagination rollouts sampled around these off-policy states provide Q-learning with additional information about alternative action not taken by the iLQG planner. In general, we did not find that off-policy rollouts were consistently better than on-policy rollouts across all tasks, but they did consistently produce good results. Performing off-policy rollouts with iLQG may be desirable in real-world domains, where a partially learned policy might take undesirable or dangerous actions. Further details of these experiments are provided in the appendix.



\subsection{Guided Imagination Rollouts with Fitted Dynamics}

In this section, we evaluated the performance of imagination rollouts with learned dynamics. As seen in Figure~\ref{fig:normq_fixedreacher_fitted}, we found that fitting time-varying linear models following the imagination rollout algorithm is substantially better than fitting neural network dynamics models for the tasks we considered. There is a fundamental tension between efficient learning and expressive models like neural nets. We cannot hope to learn useful neural network models with a small number of samples for complex tasks, which makes it difficult to acquire a good model with fewer samples than are necessary to acquire a good policy. While the model is trained with supervised learning, which is typically more sample efficient, it often needs to represent a more complex function (e.g. rigid body physics). However, having such expressive models is more crucial as we move to improve model accuracy. Figure~\ref{fig:normq_fixedreacher_fitted} presents results that compare fitted neural network models with the true dynamics when combined with imagination rollouts. These results indicate that the learned neural network models negate the benefits of imagination rollouts on our domains.


To evaluate imagination rollouts with fitted time-varying linear dynamics, we chose single-target variants of two of the manipulation tasks: the reacher and the gripper task. The results are shown in Figure~\ref{fig:normq_fixedreacher_fitted} and \ref{fig:normq_fixedgrip_fitted}. We found that imagination rollouts of length 5 to 10 were sufficient for these tasks to achieve significant improvement over the fully model-free variant of NAF.

Adding imagination rollouts in these domains provided 2-5 factors of improvement in data efficiency. In order to retain the benefit of model-free learning and allow the policy to continue improving once it exceeds the quality possible under the learned model, we switch off the imagination rollouts after 130 episodes (20,000 steps) on the gripper domain. This produces a small transient drop in the performance of the policy, but the results quickly improve again. Switching off the imagination rollouts also ensures that Q-learning does not diverge after it reaches good values, as were often observed in the gripper. This suggests that imagination rollouts, in contrast to off-policy exploration discussed in the previous section, is an effective method for bootstrapping model-free deep RL.  

It should be noted that, although time-varying linear models combined with imagination rollouts provide a substantial boost in sample efficiency, this improvement is provided at some cost in generality, since effective fitting of time-varying linear models requires relatively small initial state distributions. With more complex initial state distributions, we might cluster the trajectories and fit multiple models to account for different modes. Extending the benefits of time-varying linear models to less restrictive settings is a promising direction and build on prior work~\cite{levine2015end,fu2015one}. That said, our results show that imagination rollouts are a very promising approach to accelerating model-free learning when combined with the right kind of dynamics model.


\section{Discussion}

In this paper, we explored several methods for improving the sample efficiency of model-free deep reinforcement learning. We first propose a method for applying standard Q-learning methods to high-dimensional, continuous domains, using the normalized advantage function (NAF) representation. This allows us to simplify the more standard actor-critic style algorithms, while preserving the benefits of nonlinear value function approximation, and allows us to employ a simple and effective adaptive exploration method. We show that, in comparison to recently proposed deep actor-critic algorithms, our method tends to learn faster and acquires more accurate policies. We further explore how model-free RL can be accelerated by incorporating learned models, without sacrificing the optimality of the policy in the face of imperfect model learning. We show that, although Q-learning can incorporate off-policy experience, learning primarily from off-policy exploration (via model-based planning) only rarely improves the overall sample efficiency of the algorithm. We postulate that this caused by the need to observe both successful and unsuccessful actions, in order to obtain an accurate estimate of the Q-function. We demonstrate that an alternative method based on synthetic on-policy rollouts achieves substantially improved sample complexity, but only when the model learning algorithm is chosen carefully. We demonstrate that training neural network models does not provide substantive improvement in our domains, but simple iteratively refitted time-varying linear models do provide substantial improvement on domains where they can be applied.

\newpage
\section*{Acknowledgement}
We thank Nicholas Heess for helpful discussion and Tom Erez, Yuval Tassa, Vincent Vanhoucke, and the Google Brain and DeepMind teams for their support. 

\bibliography{arxiv1}
\bibliographystyle{icml2016}

\clearpage

\clearpage
\section{Appendix}

\subsection{iLQG}
The iLQG algorithm optimizes trajectories by iteratively constructing locally optimal linear feedback controllers under a local linearization of the dynamics $p(\x_{t+1}|\x_t,\uu_t)=\mathcal{N}(\f_{\x t}\x_t + \f_{\uu t}\uu_t, \F_t)$ and a quadratic expansion of the rewards $r(\x_t,\uu_t)$ ~\cite{tassa2012synthesis}. Under linear dynamics and quadratic rewards, the action-value function $Q(\x_t, \uu_t)$ and value function $V(\x_t)$ are locally quadratic and can be computed by dynamics programming.\footnote{While standard iLQG notation denotes $Q,V$ as discounted sum of costs, we denote them as sum of rewards to make them consistent with the rest of the paper}
  \begin{equation} \label{eq:ilqg}
  \begin{split}
 Q_{\x\uu,\x\uu t} &=  r_{\x\uu,\x\uu t} + \f^T_{\x\uu t} V_{\x,\x t+1} \f_{\x\uu t}\\
 Q_{\x\uu t} &= r_{\x\uu t} + \f_{\x\uu t}^T V_{\x,\x t+1}\\
 V_{\x,\x t} &= Q_{\x,\x t} - Q_{\uu,\x t}^T Q_{\uu,\uu t}^{-1} Q_{\uu,\x t}\\
  V_{\x t} &= Q_{\x t} - Q_{\uu,\x t}^T Q_{\uu,\uu t}^{-1} Q_{\uu t}\\
  Q_{\x,\x T} &= V_{\x,\x T} =r_{\x,\x T} \\
  \end{split}
  \end{equation}
The time-varying linear feedback controller $\bm{g}(\x_t)=\hat{\uu}_t + \kk_t + \K_t(\x_t-\hat{\x}_t) $ maximizes the locally quadratic $Q$, where $\kk_t=-Q_{\uu,\uu t}^{-1}Q_{\uu t}$, $\K_t=-Q_{\uu,\uu t}^{-1}Q_{\uu,\x t}$, and $\hat{\x}_t,\hat{\uu}_t$ denote states and actions of the current trajectory around which the partial derivatives are computed. Employing the maximum entropy objective~\cite{levine2013guided}, we can also construct a linear-Gaussian controller, where $c$ is a scalar to adjust for arbitrary scaling of the reward magnitudes,
  \begin{equation} \label{eq:ilqg_controller2}
  \begin{split}
\pi^{iLQG}_t(\uu_t|\x_t) = \mathcal{N}(\hat{\uu}_t + \kk_t + \K_t(\x_t-\hat{\x}_t), -cQ_{\uu,\uu t}^{-1})
  \end{split}
  \end{equation}
  When the dynamics are not known, a particularly effective way to use iLQG is to combine it with learned time-varying linear models $\hat{p}(\x_{t+1}|\x_t,\uu_t)$. In this variant of the algorithm, trajectories are sampled from the controller in Equation~(\ref{eq:ilqg_controller2}) and used to fit time-varying linear dynamics with linear regression. These dynamics are then used with iLQG to obtain a new controller, typically using a KL-divergence constraint to enforce a trust region, so that the new controller doesn't deviate too much from the region in which the samples were generated \cite{levine2014learning}.

\subsection{Locally-Invariant Exploration for Normalized Advantage Functions}
\label{sec:local_exp}

Exploration is an essential component of reinforcement learning algorithms. The simplest and most common type of exploration involves randomizing the actions according to some distribution, either by taking random actions with some probability \cite{mnih2015human}, or adding Gaussian noise in continuous action spaces \cite{SchulmanLAJM15}. However, choosing the magnitude of the random exploration noise can be difficult, particularly in high-dimensional domains where different action dimensions require very different exploration scales. Furthermore, independent (spherical) Gaussian noise may be inappropriate for tasks where the optimal behavior requires correlation between action dimensions, as for example in the case of the swimming snake described in our experiments, which must coordinate the motion of different body joints to produce a synchronized undulating gait.


The NAF provides us with a simple and natural avenue to obtain an adaptive exploration strategy, analogously to Boltzmann exploration. The idea is to use the matrix in the quadratic component of the advantage function as the precision for a Gaussian action distribution. This naturally causes the policy to become more deterministic along directions where the advantage function varies steeply, and more random along directions where it is flat. The corresponding policy is given by
\begin{equation} \label{eq:normq_controller}
\begin{split}
\pi(\uu|\x) &= \exp^{Q(\x,\uu|\theta^Q)} /\int \exp^{Q(\x,\uu|\theta^Q)} d \uu\\
&= \mathcal{N}(\mmu(\x|\theta^\mu), c\PP(\x|\theta^P)^{-1}).
\end{split}
\end{equation}
Previous work also noted that generating Gaussian exploration noise independently for each time step was not well-suited for many continuous control tasks, particularly simulated robotic tasks where the actions correspond to torques or velocities \cite{lillicrap2015continuous}. The intuition is that, as the length of the time-step decreases, temporally independent Gaussian exploration will cancel out between time steps. Instead, prior work proposed to sample noise from an Ornstein-Uhlenbech (OU) process to generate a temporally correlated noise sequence \cite{lillicrap2015continuous}. We adopt the same approach in our work, but sample the innovations for the OU process from the Gaussian distribution in Equation~\ref{eq:normq_controller}. Lastly, we note that the overall scale of $\PP(\x|\theta^P)$ could vary significantly through the learning, and depends on the magnitude of the cost, which introduces an undesirable additional degree of freedom. We therefore use a heuristic adaptive-scaling trick to stabilize the noise magnitudes.

\begin{figure}
	\centering
	\includegraphics[width=0.99\linewidth]{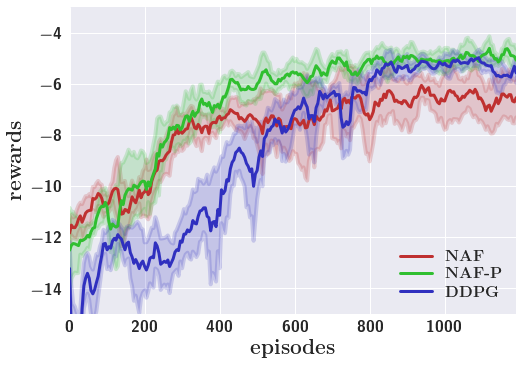}
	\caption{NAF with exploration noise generated using the precision term (NAF-P) slightly outperforms the best DDPG result. Precision term is not used until step 50,000.}
	\label{fig:normq_canada2d}
\end{figure}
Using the learned precision as the noise covariance for exploration allowed for convergence to a better policy on the ``canada2d'' task, which requires using an arm to strike a puck toward a target, as shown in Figure~\ref{fig:normq_canada2d}, but did not make a significant difference on the other domains.
%
%
%
 \subsection{Q-Learning as Variational Inference}
 \label{sec:qkl}
 NAF, with our choice of parametrization, can only fit a locally quadratic Q-function. To understand its implications and expected behaviors, one approach is to approximately view the Q-learning objective as minimizing the exclusive Kullback-Leibler divergence ($KL$) between policy distributions induced by the fitted normalized Q-function $\hat{Q}$ and the true Q-function $Q$. This can be derived by assuming (1) no bootstrapping and the exact target $Q$ is provided per step, (2) no additive exploration noise, i.e. fully on-policy, and (3) use KL loss instead of the least-square. Specifically, let $\pi$ and $\hat{\pi}$ be corresponding policies of $Q$ and $\hat{Q}$ respectively, an alternative form of Q-learning could be optimizing the following objective: 
 \begin{equation} \label{eq:qvi}
 \begin{split}
 L_e(\hat{Q}) =  \mathbb{E}_{\x_t\sim\rho^{\hat{\pi}}}[\tilde{KL}(\hat{\pi}||\pi)] = \mathbb{E}_{\x_t\sim\rho^{\hat{\pi}},\uu_t\sim\hat{\pi}}[\hat{Q} - Q] \\
 \end{split}
 \end{equation} 
 We can thus intuitively interpret NAF as doing variational inference to fit a Gaussian to a distribution, and it has mode-seeking behavior. Empirically such behavior enables NAF to learn smoother and more precise controllers, as most effectively illustrated by three-joint reacher and peg insertion experiments, and substantial improvements in terms of convergence speeds in many other representative domains explored in the main paper. 
 
 %


\subsection{Descriptions of Task Domains}
\label{sec:domains}
Table~\ref{tab:domains} describes  the task domains used in the experiments.

\subsection{More Results on Normalized Advantage Functions}
\label{sec:exp_normq_more}
\begin{figure*}[t!]
\begin{subfigure}[t]{0.33\textwidth}
	\centering\captionsetup{width=.8\linewidth}
	\includegraphics[width=0.99\linewidth]{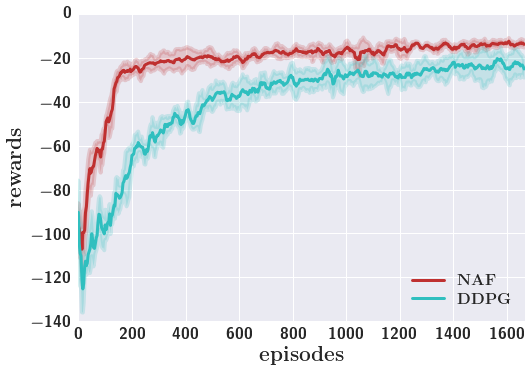}
	\caption{NAF significantly outperforms DDPG on moving gripper.}
	\label{fig:normq_movinggripper}
\end{subfigure}
\begin{subfigure}[t]{0.33\textwidth}
	\centering\captionsetup{width=.8\linewidth}
	\includegraphics[width=0.99\linewidth]{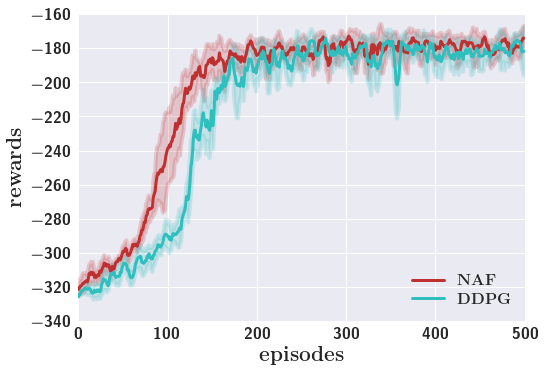}
	\caption{NAF converges faster than DDPG on swimmer6. }
	\label{fig:normq_swimmer6}
\end{subfigure}
\begin{subfigure}[t]{0.33\textwidth}
	\centering\captionsetup{width=.8\linewidth}
	\includegraphics[width=0.99\linewidth]{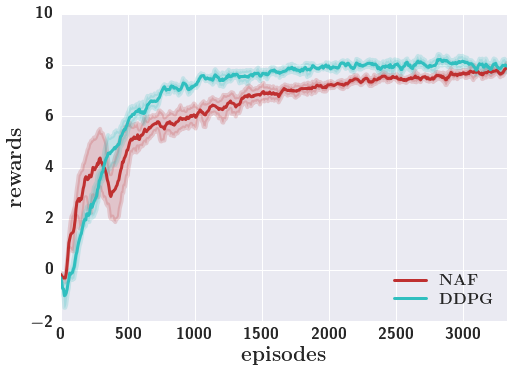}
	\caption{DDPG converges faster than NAF on cheetah. }
	\label{fig:normq_cheetah}
\end{subfigure}
\caption{NAF vs DDPG on three domains.}
\label{fig:ddpg_naf}
\end{figure*}

Figures~\ref{fig:normq_movinggripper},~\ref{fig:normq_swimmer6}, and~\ref{fig:normq_cheetah} provide additional results on the comparison experiments between DDPG and NAF. As shown in the main paper, NAF generally outperforms DDPG. In certain tasks that require precision, such as peg insertion, the difference is very noticeable. However, there are also few cases where NAF underperforms DDPG. The most consistent of such cases is cheetah. While both DDPG and NAF enable cheetah to run decent distances, it is often observed that the cheetah movements learned in NAF are little less natural than those from DDPG. We speculate such behaviors come from the mode-seeking behavior that is explained in Section~\ref{sec:qkl}, and thus exploring other parametric forms of NAF, such as multi-modal variants, is a promising avenue for future work.

\subsection{More Results on Evaluating Best-Case Model-Based Improvement with True Models}
\begin{figure*}[t!]
	 \begin{subfigure}[t]{0.33\textwidth}
	 	\centering\captionsetup{width=.8\linewidth}%
	 	\includegraphics[width=0.99\linewidth]{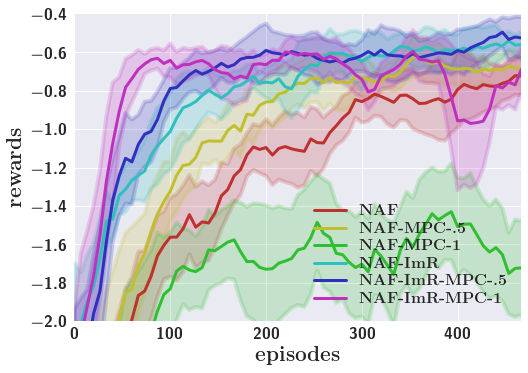}
	 	\caption{NAF on multi-target reacher. Insignificant gain from mixing MPC experience. Significant gain from imagination rollouts.}
	 	\label{fig:fic15_reacher}
	 \end{subfigure}
\begin{subfigure}[t]{0.33\textwidth}
	\centering\captionsetup{width=.8\linewidth}
	\includegraphics[width=0.99\linewidth]{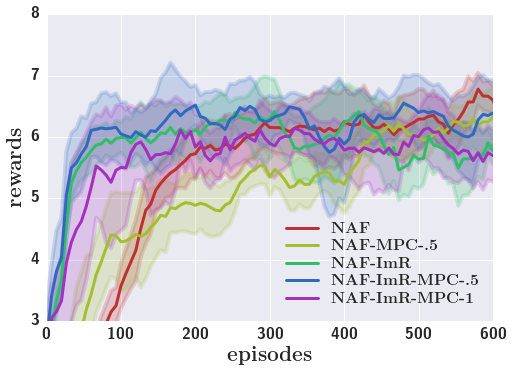}
	\caption{NAF on cheetah. Great speeds up with imagination rollouts, no gain from mixing MPC experiences.}
	\label{fig:normq_cheetah_fic}
\end{subfigure}
\begin{subfigure}[t]{0.33\textwidth}
	\centering\captionsetup{width=.8\linewidth}
	\includegraphics[width=0.99\linewidth]{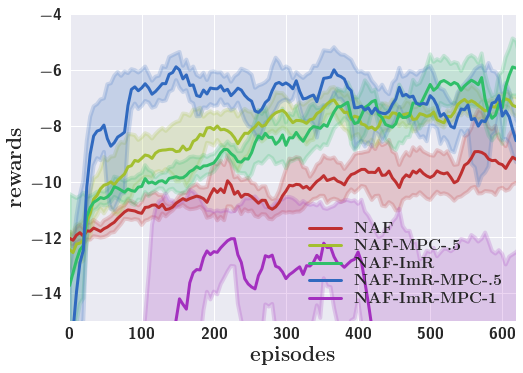}
	\caption{NAF on canada2d. Very significant speed-ups from mixing MPC experiences, both with or without the rollouts.}
	\label{fig:normq_canada2d_fic}
\end{subfigure}
	\caption{NAF on multi-target reacher, cheetah, and canada2d, with model-based acceleration using true dynamics: ``ImR" denotes using the imagination rollout, $l=10$ steps. ``MPC-$x$" indicates mixing $x$ fraction of MPC episodes.\vspace{-0.1 in}}
	\label{fig:naf_imr}
\end{figure*}

 \begin{table}[ht]
 	\centering 
 	\footnotesize
 	\begin{tabular}{|c  |c |c |c| c| c  | }
 		\hline
 		Domains	&-	& 0.5 &ImR&ImR,0.5&ImR,1 \\
 		
 		\hline
 		Reacher &-0.488& -0.449 & -0.448 & \textbf{-0.426}& -0.548 \\
 		episodes&740 & 670 &450 &430 &\textbf{90}\\
 		\hline
 		Canada2d &-6.23& -6.23 & -5.89 & \textbf{-5.88} & -12.0\\
 		episodes& 1970&1580 &570 &\textbf{140} & 210\\
 		\hline
 		Cheetah &7.00& 7.10 & \textbf{7.36} & 7.29 &6.43   \\	
 		episodes& 580& 1080 & 590& 740&\textbf{390}\\
 		\hline
 	\end{tabular}
 	\caption{\footnotesize Best-case model-based acceleration with true dynamics models. Best test rewards of NAF policies (first row), and the episodes it required to reach 5\% of the best value (second row). ``0.5" and ``1" correspond to the fraction of MPC episodes. ``ImR" means using imagination rollout with rollout length $l=10$ for reacher, canada2d, and $l=5$ for cheetah.\vspace{-0.1 in}}
 	\label{tab:mixq}
 \end{table}
In the main paper, iLQG with true dynamics is used to generate guided exploration trajectories. While iLQG works for simple manipulation tasks with small number of initial states, it does not work well for random target reacher or complex locomotion tasks such as cheetah. We therefore run iLQG in model-predictive control (MPC) mode for the experiments reported in Figures~\ref{fig:normq_canada2d_fic},~\ref{fig:normq_cheetah_fic}, and~\ref{fig:fic15_reacher}, and Table~\ref{tab:mixq}. It is important to note that for those experiments, the hyper-parameters were fixed (batch normalization is on, learning rate is $10^{-3}$, and exploration size is 0.3), and thus the results differ slightly from the experiments in the previous section.

In cheetah and other complex locomotion tasks, MPC policy is usually sub-optimal, and thus poor performance of mixing MPC experience in Figure~\ref{fig:normq_cheetah_fic} is expected. On the other hand, MPC policy works reasonably in hard manipulation tasks such as canada2d, and there is significant gain from mixing MPC experience as Figure~\ref{fig:normq_canada2d_fic} shows. However, the most consistent gain comes from using imagination rollouts in all three domains. In particular, Figure~\ref{fig:normq_canada2d_fic} shows that in canada2d, MPC experiences gives very good trajectories, i.e. those that hit balls in roughly the right directions, and doing rollouts can generate more of this useful experience, enabling canada2d to learn very quickly. While with true dynamics having the imagination experience directly means more experience and such result may be trivial, it is still important to see the benefits of rollouts which only explore up to $l=10$ steps away from the real experience, as reported here. This is an interesting result, since this means the dynamics model only needs to be accurate around the data trajectories and this significantly lessens the requirement on fitted models. 

\begin{table*}[ht]
	\centering 
	\footnotesize
	\begin{tabular}{C{1.5cm}  |L{6cm} ||C{1.5cm}  |L{6cm} }
		
		Domain	&Description	& Domain & Description\\
		
		\hline \
		Cartpole & The  classic  cart-pole  swing-up  task.   Agent  must  balance  a  pole  attached to a cart by applying forces to the cart alone.  The pole starts
		each episode hanging upside-down. & Reacher & Agent is required to move a 3-DOF arm from random starting locations
		to random target positions.  \\ \hline \
		Peg & Agent is required to insert the tip of a 3-DOF arm from locally-perturbed starting locations to a fixed hole.&
		Gripper & Agent must use an arm with gripper appendage to grasp an object and
		manuver the object to a fixed target.\\ \hline \
		GripperM & Agent must use an arm with gripper attached to a moveable platform to
		grasp an object and move it to a fixed target.&
		Canada2d & Agent is required to use an arm with hockey-stick like appendage to hit
		a ball initialzed to a random start location to a random target location.\\ \hline \
		Cheetah & Agent should move forward as quickly as possible with a cheetah-
		like body that is constrained to the plane. &
		Swimmer6 & Agent should swim in snake-like manner toward the fixed target using six joints, starting from random poses.\\ \hline \
		Ant & The four-legged ant should move toward the fixed target from a fixed starting position and posture.&
		Walker2d & Agent should move forward as quickly as possible with a bipedal walker constrained to the plane without falling down or pitching the torso too
		far forward or backward. 
	\end{tabular}
	\caption{List of domains. All the domains except ant are 2D.\vspace{-0.1 in}}
	\label{tab:domains}
\end{table*}

\end{document}